%% file: main.tex
\documentclass[11pt,a4paper]{article}
\usepackage[hyperref]{acl2020}
\usepackage{times}
\usepackage{latexsym}

\usepackage{microtype}

\usepackage{tikz}

\usepackage{soul}
\usepackage{url}
\usepackage[hidelinks]{}
\usepackage[utf8]{inputenc}
\usepackage{caption}
\usepackage{graphicx}
\usepackage{amsmath}
\usepackage{booktabs}
\usepackage{tikz}
\tikzset{every picture/.style={remember picture}}
\usetikzlibrary{tikzmark}
\usepackage{multirow}
\usepackage{makecell}
\usepackage{hhline}
\usepackage{paralist}
\usepackage{array}
\usepackage{fancyhdr}
\usepackage{tabu}
\usepackage[normalem]{ulem}

\usepackage{calc}
\newlength\myheight
\newlength\mydepth
\settototalheight\myheight{Xygp}
\newcommand*\inlinegraphics[1]{%
  \settototalheight\myheight{Xygp}%
  \settodepth\mydepth{Xygp}%
  \raisebox{-\mydepth}{\includegraphics[height=\myheight]{#1}}%
}
\newcommand{\oursys}{XFBoost}
\usepackage{amssymb,amsmath,amsthm,enumitem}

\usepackage[ruled,vlined]{algorithm2e}

\title{XFBoost: Improving Text Generation with Controllable Decoders}
\author{Xiangyu Peng \\
  Georgia Institute of Technology \\
  \texttt{xpeng62@gatech.edu} \\\And
  Michael Sollami\\
  Salesforce Einstein \\
  \texttt{msollami@salesforce.com} \\}

\date{}

\usepackage{xcolor}

\newcommand{\mantis}{MAnTiS}
\begin{document}
\maketitle

\begin{abstract}
Multimodal conditionality in transformer-based natural language models has demonstrated state-of-the-art performance in the task of product description generation. Recent approaches condition a language model on one or more images and other textual metadata to achieve near-human performance for describing products from e-commerce stores~\cite{jain2021multimodal}. However, generated descriptions may exhibit degrees of inaccuracy or even contradictory claims relative to the inputs of a given product. In this paper, we propose a controllable language generation framework called \textit{Extract-Finetune-Boost} (\oursys), which addresses the problem of inaccurate low-quality inference. By using visual semantic attributes as constraints at the decoding stage of the generation process and finetuning the language model with policy gradient techniques, the \oursys{} framework is found to produce significantly more descriptive text with higher image relevancy, outperforming baselines and lowering the frequency of factually inaccurate descriptions. We further demonstrate the application of \oursys{} to online learning wherein human-in-the-loop critics improve language models with active feedback.
\end{abstract}

\begin{figure}[t!]
    \centering
    \includegraphics[width=\linewidth]{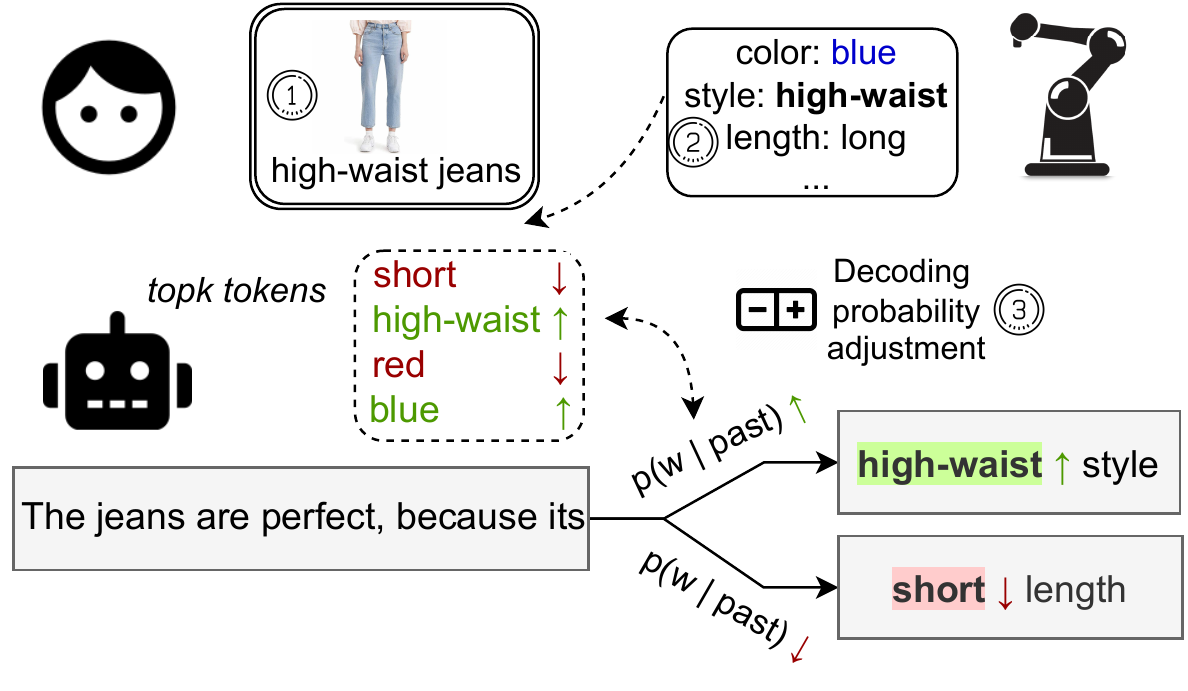}
    \caption{An overview of the \oursys{} system.
    (1) A set of product images and product title start the description generation process.
    (2) Attributes of the product are extracted by attribute extraction model \inlinegraphics{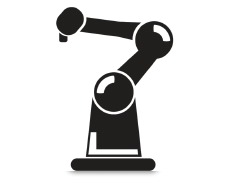}.
    (3) At the decoding stage, sampling probability of the topk tokens are adjusted by its similarity with extracted attributes. Our system guides generations towards those that satisfy the extracted attributes and given constraints.
    }
    \label{fig:figure_1}
\end{figure}

\section{Introduction}
\label{sec:intro}


Multimodal models, visolinguistic transformers in particular, have demonstrated state-of-the-art performance in vision-and-language tasks such as captioning \cite{chen2015microsoft}, visual question answering \cite{antol2015vqa}, and visual reasoning \cite{suhr2018corpus}. 
A natural extension to the captioning task is conditional Natural Language Generation (NLG).
In this setting, multimodal NLG models are prefixed on image and associated textual data and trained to produce natural language descriptions conditioned on both image and text input. The description generation task differs from the captioning task in that text produced requires not only the accuracy of describing images coherently, but also has to satisfy complex lexical constraints.

Large-scale, transformer-based neural language models---such as ELMo \cite{elmo}, BERT \cite{BERT}, and  GPT \cite{radford2019language}---are widely used in state-of-the-art multimodal models.  They are first pretrained on vast corpora of internet scraped text and books, then finetuned on supervised data together with a task-related head. These language models generate text that is statistically representative of the corpora on which they were trained.
Consequently, it is often possible for language models to generate low-quality descriptions, i.e. containing contradictory or irrelevant items, for out-of-distribution images not similar to those seen during training.

\paragraph{Overview}
Inspired by \citeauthor{jain2021multimodal} \shortcite{jain2021multimodal}, our paper focuses on improving the \textit{quality of generated product descriptions}---describing images accurately with common-sense and human-preferred language that conform to conditional inputs (such as brand, title, or category). In this paper, we propose a simple but efficient controllable language generation framework (Figure~\ref{fig:figure_1}) called E\textbf{x}tract-\textbf{F}inetune-\textbf{Boost} (\oursys) which addresses this issue of output incoherence in multimodal NLG models.
\oursys{} integrates Multimodal Adaptation for Text Synthesis (\mantis) \cite{jain2021multimodal}---a general approach for multimodal conditionality in transformer-based NLG models---for natural language generation.

An attribute extraction model (\textit{Extract}) is first trained to produce lexical constraints on the generation whose satisfaction we enforce (\textit{Boost}) by controlling the decoding stage of sequence generation. We show that this approach improves the probability of correct attributes during inference.
A reward model is constructed to predict qualitative scores on human-labeled sets of the generated descriptions. Reinforcement techniques are then employed to \textit{finetune} the language model with the derived reward model. \oursys{} guides the multimodal language model to produce natural language descriptions of images with higher quality at the decoding stage and forces the model to produce more human preferred language through the finetuning stage.

Our contributions are two-fold: (1) we propose a controllable language generation framework, \oursys, which controls the
decoding stage of sequence generation to substantially improve the quality of generated description and then finetune language model with human-preference; and (2) we conduct a thorough experimental study against the strong baseline that shows that \oursys{} produces more accurate descriptions while maintaining desired attributes of prose attractiveness and fluency. We apply reinforcement learning techniques to further finetune the model. The \oursys{} method of controlling decoding successfully generalizes to any transformer-based language model.

\section{Related Work}
\label{sec:related}
Multimodal models \cite{yang2016stacked, anderson2018bottom, jiang2018pythia, li2019visualbert, lu2019vilbert, jain2021multimodal} have been proved to successfully bridge vision and language.
These approaches often consist of a text encoder, an image feature extractor, and a multimodal fusion module. 
Many multimodal models focus on common natural language generation tasks such as captioning \cite{xu2015show, anderson2018bottom, herdade2019image, yang2020fashion} or description with inference \cite{ziegler2019encoder, jain2021multimodal}.

\paragraph{Conditional multimodal language model.}
Models conditioned with pretrained language include noisy channel modeling \cite{yee2019simple} and fusion approaches \cite{gulcehre2015using, sriram2017cold} that concatenate hidden states of the conditional model with that of language model to predict the next word. 
\citeauthor{ziegler2019encoder} \shortcite{ziegler2019encoder} and \citeauthor{jain2021multimodal} \shortcite{ziegler2019encoder} considered modality conditioning by passing inputs from each modality through modality-specific encoders. We take the multimodal language model, \mantis, by \citeauthor{jain2021multimodal}\shortcite{jain2021multimodal} as baseline and conduct a thorough experimental study on our framework with \mantis.

\paragraph{Guiding generation by controlling decoding stage.}
\citeauthor{krause2020gedi}\shortcite{krause2020gedi} proposed generative discriminator guided
sequence generation (GeDi)---guiding generation at each step by computing classification probabilities for all possible next tokens via Bayes rule by normalizing over two class-conditional distributions.
\citeauthor{lin2021plug}\shortcite{lin2021plug} controls the generation towards given continuous-weighted control codes. 
\citeauthor{lu2020neurologic}\shortcite{lu2020neurologic} proposed Neurologic Decoding---an efficient and general method for generating with arbitrary predicate logic constraints.

\paragraph{Finetuning language models with reinforcement learning.}
The Plug and Play Language Models (PPLM) \cite{dathathri2019plug} apply attribute classifiers to finetune language models; 
the technique is demonstrated via generating text with a target sentiment and also decreasing the frequency of toxic language.
\citet{ziegler2019fine} use a reinforcement learning method on GPT-2 to favor human-preferred text. 
\citeauthor{peng2020reducing}\shortcite{peng2020reducing} finetuned language model using reinforcement learning techniques and a normative text classifier to produce reward  and  punishment values.
\citeauthor{peng2021inferring}\shortcite{peng2021inferring} finetuned language model with common-sense inference and reinforcement.
Similar with these approaches, we also trained a reward model to predict the human preference of the generated descriptions. And then finetune language model with the help of reward model.

\section{Methods}
\label{sec:methods}
\begin{figure*}[t]
    \centering
    \includegraphics[width=\textwidth]{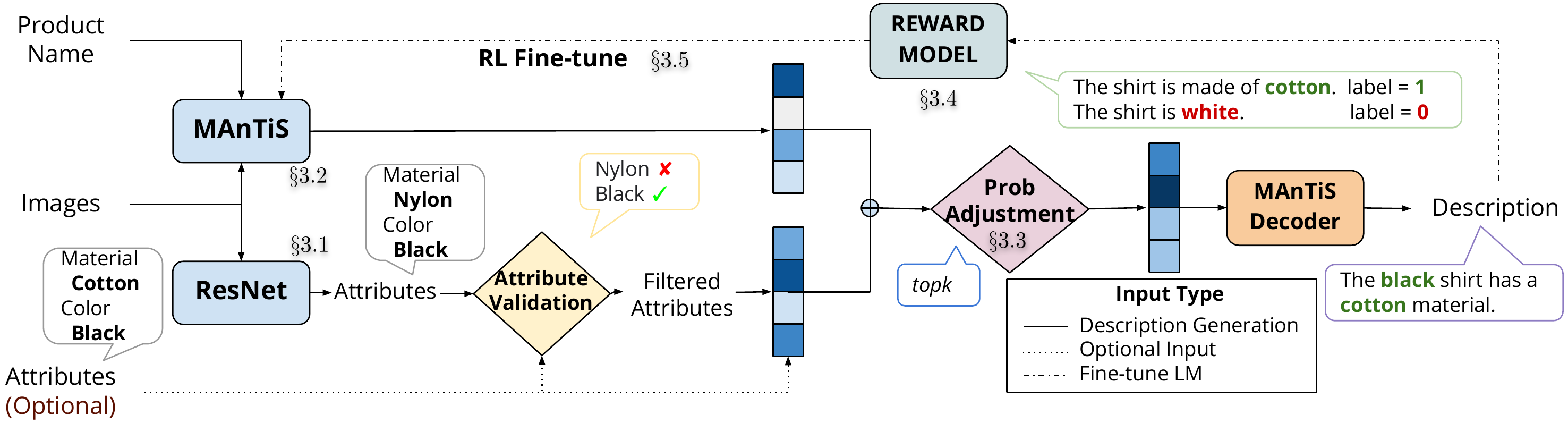}
    \caption{The \oursys{} system for generating improved product descriptions.}
    \label{fig:pipeline}
\end{figure*}

In the conventional setup for the text description problem, we are given a product image $m$ and a product natural language title $t$, and the task is to generate a product description $s$. A proposed framework Extract-Finetune-Boost (\oursys) is a procedure for optimizing a given language model pretrained for this task (an overview is given in Figure~\ref{fig:pipeline}). Generally, \oursys{} generates a product description $s$ by performing the following steps:
\begin{enumerate}[noitemsep,topsep=3pt,itemsep=3pt]
    \item Product attributes $a_1, ..., a_n$ are extracted by an visual attribute extractor. The training of the attribute extractor is shown in Section~\ref{sec:ae}.
    \item We condition a multimodal language model (See Section~\ref{sec:multi}) on the product image $m$ and product natural language title $t$.
    \item Description $s$ is generated token by token from the multimodal language model. Probability of the top-k tokens in the $i$th position at decoding stage, $p_\theta(x_{i} \mid x_{<i})$, are modified by its similarity with attributes $a_1, ..., a_n$, which is illustrated in Section~\ref{sec:boosting}. 
    \item We sample next token, $x_i$, with the modified probability $p'_\theta(x_{i} \mid x_{<i})$, see Equation~\ref{eq:modified_equation}.
    \item Repeat steps 3 and 4 until \texttt{<eos>} token is sampled.
    \item (Optional) Reward model evaluates whether $s$ are preferred by humans. Reward Model architecture is introduced in Section~\ref{sec:reward}. If not, Repeat steps 3 to 6 until $s$ is predicted as human-preferred by reward model.
\end{enumerate}

The system is simple but effective---\oursys{} evaluates whether next possible tokens are preferred based on the extracted attributes. Unlike models trained or finetuned on image captions, \oursys{} boosts the sampling probability of the tokens which are highly relevant with extracted attributes. 
\oursys{} is prone to generate a description which has less contradiction with the image.

After generating descriptions, \oursys{} keeps improving the language model by:
\begin{enumerate}[noitemsep,topsep=3pt,itemsep=3pt]
    \item Human turkers give their preference on each generated description, given the image and product title.
    \item Human preference are used to train a reward model, which is used to filter out the generated descriptions in generation process and also produce reward and punishment values for finetuning language model (See Section~\ref{sec:reward}).
    \item Generated description and reward model are used to finetune the language model (See Section~\ref{sec:finetune}).
\end{enumerate}

\oursys{} provides a continuous learning opportunity. With the human help, the language model and reward model are improved every time, making the language model adapt to different language style and image types.

\subsection{Attributes Extraction}
\label{sec:ae}

Salient visual attributes are important factors in the description generation, and so image attributes are extracted first to help provide lexical constraints of generation. 
We train a ResNet-50 model\cite{he2016deep} to predict attributes of products from the apparel industry, based on a catalog of product images, namely the Kaggle iMaterialist Challenge (Fashion) at FCVC5. 

Attributes have $N$ classes and for each class $k$, the classification model will give a prediction $a_{k} \in A_{k}$, where $k \in [1, N]$ and $A_{k} = \{a_{k,1}, a_{k,2}, ...\}$ is one attribute class.
Hence, for each image, the classification model will provide a prediction, $A = [a_1, ..., a_{N}]$. 
We also consider cases when ground-truth attributes---$\hat{A} = \{\hat{a_1}, \hat{a_2}, ...\}$ where $\hat{a_k} \in \hat{A_k}$ and $\hat{A_k}$ are attributes classes---are provided during description generation. 
We first create an \textit{attribute vocabulary}---includes attribute classes $\widetilde{A_{k}}$ with multiple attributes $\widetilde{a_{kj}} \in \widetilde{A_{k}}$. For example, in the class ``material'', we have over $50$ different material attributes, such as ``cotton'', ``nylon'' and ``polyester''.
Then we conduct \textit{attribute validation} between extracted attributes $A$ and ground-truth attributes $\hat{A}$.

\begin{algorithm}[t] 
\SetAlgoLined
\KwData{$V()$, A, $\hat{A}$}
\KwResult{$A' \subset A$}
$A' \leftarrow A$\;
\For{$k \gets 1$ to $N$}{ 
    \eIf{$\hat{a_k} == \emptyset$ or $V(a_k) == V(\hat{a_k})$}{
        keep $a_k$ in $A'$\;
        }{remove $a_k$ from $A'$\;}
    }
\caption{Attribute Validation}
\label{alg:attr-validation}
\end{algorithm}

In Algorithm~\ref{alg:attr-validation}, $V()$ is a function, which maps attributes to the word in attribute vocabulary. The filtered attributes $A'$ are used to produce lexical constraints of generation.


\subsection{Multimodal Language Model}
\label{sec:multi}

Conditioning a network on both the image $m$ and natural langugae title $t$, multimodal language model generates descriptive text. In this paper, following \citeauthor{jain2021multimodal} \shortcite{jain2021multimodal}, we first encode images by extracting the embedding $e_m$ form of the last fully-connected layer of a pretrained ResNet-152 model. 
We then condition the pretrained \mantis\cite{jain2021multimodal} model with image embedding $e_m$ and title $t$, where $e_m$ is regarded as a single dense token per image
whose dimension depends on the ResNet model.
Lastly, we train the multimodal model by minimize the cross-entropy loss given by \cite{radford2019language}:
\begin{align}
loss(X, y) = &-\log\left(\frac{\exp(X[y])}{\sum_{i \in V} \exp(X[i])}\right) 
\label{eq:loss_cross}
\end{align}
where $X$ is a vector containing output logits
and \textit{y} is the index of the word from the ground truth in $X$.
$V$ is the model's vocabulary.

\subsection{Decoding Probability Adjustment}
\label{sec:boosting}

Transformer-based language model generates sequences by computing joint probabilities over symbols as the product of conditional probabilities \cite{jelinek1980interpolated, bengio2003neural, radford2019language}.

\begin{align}
    p_\theta(x_{1:n}) &=\prod_{i=1}^{n} p_\theta\left(x_{i} \mid x_{1}, \ldots, x_{i-1}\right) \nonumber \\
    &=\prod_{x=1}^{n} p_\theta(x_{i} \mid x_{<i})
\end{align}

To encourage the language model to (1) generate more tokens which are highly relevant with extracted attributes $A' = [a_1, ..., a_{n'}]$ in Equation~\ref{alg:attr-validation} and (2) punish those tokens contracting with images, we modify the conditional probability $p_\theta(x_{i} \mid x_{<i})$ as follows:

\begin{align}
    p'_\theta(\vec{x_i} \mid x_{<i}) &=p_\theta(\vec{x_i} \mid x_{<i}, A') \nonumber \\
    &= \delta(\vec{x_i}) \times p_\theta(\vec{x_i} \mid x_{<i})
    \label{eq:modified_equation}
\end{align}
    
where $\delta(\vec{x_i}) = [\delta(x_{i1}), ..., \delta(x_{ik})]$ is a function of the relationship between top-k tokens $\vec{x_i} = [x_{i1}, x_{i2}, ... x_{ik}]$ and attributes $A$.
They serve as a punishment or reward value vector for finetuning language model.

\begin{equation}
  \delta(\vec{x_{ij}}) =
    \begin{cases}
      1 & \text{if $\sum_{k'=1}^{n'} \sigma(x_{ij}, a_{k'}) = 0$}\\
      1 + \mu & \text{if $\sum_{k'=1}^{n'} \sigma(x_{ij}, a_{k'}) > 0$}\\
      1 - \mu & \text{if $\sum_{k'=1}^{n'} \sigma(x_{ij}, a_{k'}) < 0$}
    \end{cases}       
\end{equation}

where $\mu$ is a hyper-parameter to control the strength of the penalty and $\sigma(x_{ij}, a_{k'})$ is a function of the relationship between token $x_{ij}$ and attribute $a_{k'}$. 
When the top-k token $x_{ij}$ shares a high similarity with a predicted attributes $a_{k'} \in A'_{k'}$, then the probability of this token will be boosted. 
Conversely, when $x_{ij}$ shares a high similarity with any other attributes $a_{j'} \in A'_{k'}$ in the class $k'$ in the product detail vocabulary $\widetilde{A}$, then we will punish this token because it contracts with the image. 
For example, when we extract ``\textit{cotton}'' from the images in the ``\textit{material}'' class, token ``\textit{nylon}'' will be punished and has less probability to be sampled.

\begin{align}
  \sigma(x_{ij}, a_{k'}) =
    \begin{cases}
      1 & \text{if $x_{ij}  \simeq a_{k'} \in \widetilde{A_{k'}}$}\\
      0 & \text{if  $x_{ij} \notin \widetilde{A_{k'}}$}\\
      -1 & \text{if $x_{ij}  \simeq a_{j'} \in \widetilde{A_{k'}}$, $a_{j'} \neq a_{k'}$}
    \end{cases}       
\end{align}

where $\simeq$ indicates $x_{ij}$ and $a_{k'}$ share similarity over the semantic similarity threshold.

\subsection{Reward model} 
\label{sec:reward}
In our framework, we build a reward model to predict human preference of automatically generated descriptions.
Reward model shared the similar architecture with multimodal language model in Section~\ref{sec:multi}. Given the image $m$, title $t$, attributes $A$ and generated description $S$, reward model predicts whether the description will be preferred by humans. Since GPT-2 is a decoder transformer, the last token of the input sequence is used to make predictions. 
Reward model training details can be found in Section~\ref{sec:rw_eva}.

Reward model is used to (1) filter the generated descriptions and (2) finetune language models. 
Only generated descriptions with high probability to be accepted by human which are predicted by reward model, will be used as product descriptions.
Reward model can also be used to provide the punishment or reward value for finetuning language model with reinforcement techniques (Section~\ref{sec:finetune}). 

\subsection{Finetune Language Model} 
\label{sec:finetune}
\begin{figure}[t]
    \centering
    \includegraphics[width=0.8\linewidth]{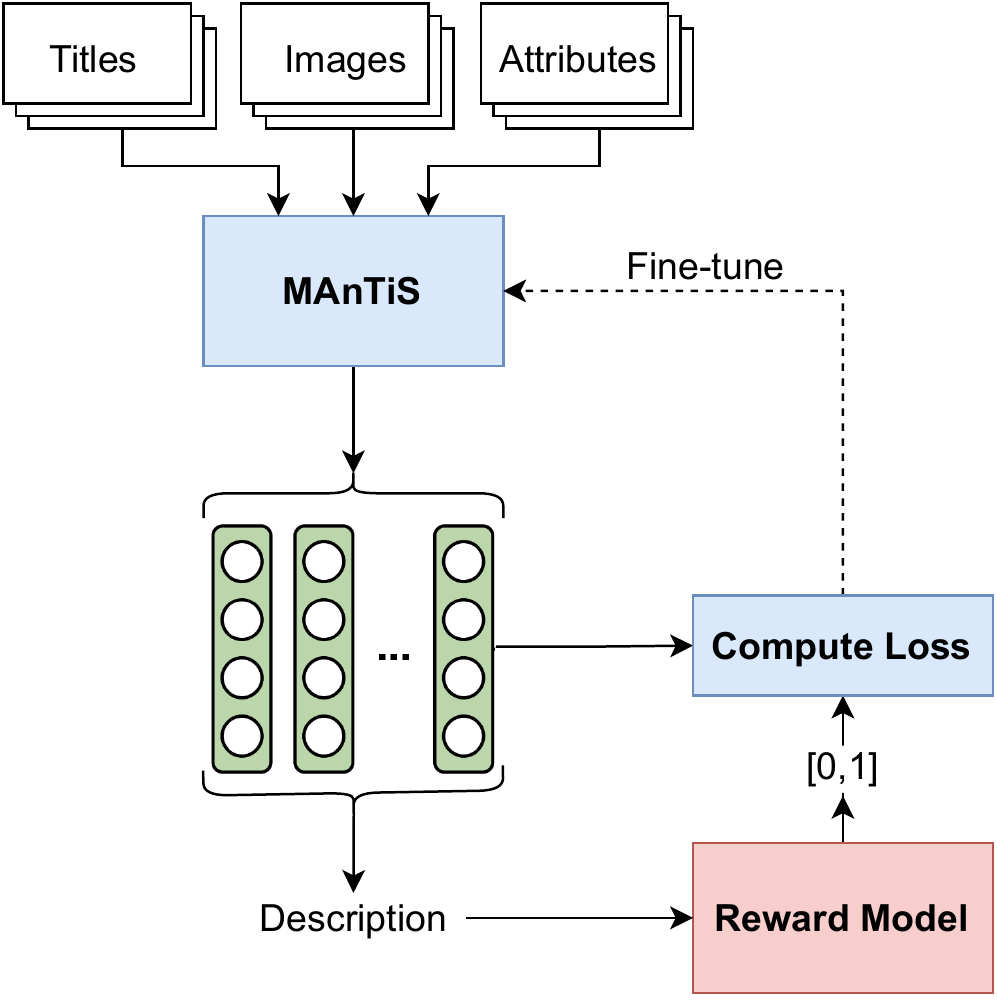}
    \caption{Pipeline for finetuning language model with reward model. Loss is back-propagated through the output logits to language model.}
    \label{fig:fine-tune}
\end{figure}

After training a reward model, multimodal language model is fine-tuned using the following policy gradient reinforcement learning technique.
To punish the generation of ``unpreferred" sentences, we modify cross-entropy loss of sentence $s$ given by \citet{radford2019language} as,


\begin{equation}
{\rm{loss}}_{{\rm{RL}}}(s)=\frac{1}{n} \sum_{j \in s}{\rm{loss}}_{{\rm{w}}}(X_{j}, y_{j}) + u(s)
\label{eq:loss_RL}
\end{equation}

where $X_j$ is the $j^{\rm th}$ logit vector, and $y_j$ is the ground truth index for the $j^{\rm th}$ word.
${\rm{loss}}_{{\rm{w}}}()$ is cross-entropy loss function.
$u(s)$ is a function of the output of the reward model converted into a {\em punishment} value.

\begin{equation}
u(s)=\rho(1-C(s))(\frac{1}{n}\sum_{j\in s}{{\rm{loss}}_{\rm w}(X_j,y_j))}
\label{eq:penalty}
\end{equation}

where $\rho$ is a constant ($\rho \geq 1$) controlling the strength of punishment---sentences which are not preferred by human participants incur more loss. $C(s)$ is the binary $\{0,1\}$ label given by the reward model. 

The finetuning process is as follows: given a set of product images, titles and attributes, multimodal language model is used to generate descriptions without adjusting sampling probability at decoding stage (\S\ref{sec:boosting}). 
These descriptions are fed through reward model, which outputs the binary label we treat as a reward.
Descriptions labeled as $0$ are those with undesirable characteristic, not preferred by human.
As per Equation~\ref{eq:penalty}, the reward is used to calculate the punishment score.  
The sentence loss is averaged to obtain the token-level loss, which is then added to each logit from the generated description and the loss is back-propagated into language model. 
The process is illustrated in Figure~\ref{fig:fine-tune}.

\section{Experiments}
\label{sec:exp}
We conducted three experiments to evaluate the performance of \oursys{}. 
The first experiment assessed whether XB (extraction and boosting in Section~\ref{sec:ae} and \ref{sec:boosting}) could guide multimodal language models to generate more accurate and interesting descriptive text.
The second evaluated the performance of reward model.
The third experiment evaluated whether our reward-based fine tuning (Section~\ref{sec:finetune}) led the language model to produce higher-quality descriptions that human critics actually preferred.

\paragraph{Datasets.}
We evaluate our models on the FAshion CAptioning Dataset (FACAD) \cite{yang2020fashion}, which contains 993K images and 130K corresponding enchanting and diverse
descriptions. 
Given a product’s name and image, the multimodal language model is trained to generate an e-commerce relevant description.

\paragraph{Baselines.}
We compare our model with Multimodal Adaptation for Text Synthesis (\mantis) \cite{jain2021multimodal}---a general approach for multimodal conditionality in transformer-based natural language generation models. 

\begin{figure}[t]
    \centering
    \includegraphics[width=\linewidth]{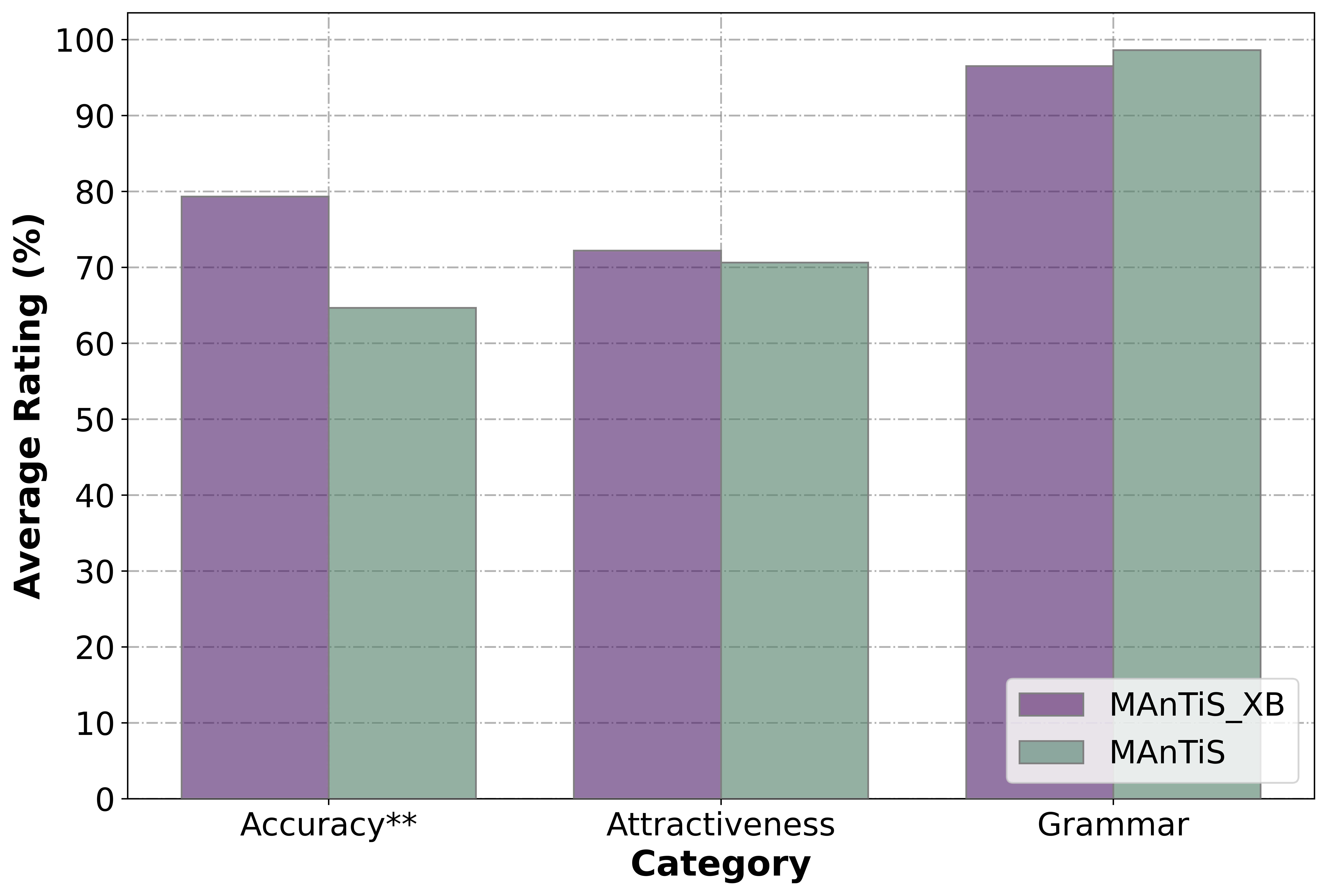}
    \caption{Human evaluation results comparing XB  with baseline, $\ast$ indicates $p < 0.05$.}
    \label{fig:eb_human}
\end{figure}

\input{tables/auto_1}
\input{tables/example_rl}
\input{tables/reward}



\subsection{Extraction and Boosting Evaluation}
\label{sec:eb_eva}
In this paper, we evaluate our model using human participant evaluation, asking a set of questions to measure three linguistic qualities that includes accuracy, attractiveness and grammar.

\begin{compactitem}
    \item \textit{Accuracy}: Is this description accurately describe the product? The main purpose of product description is to introduce the product accurately.
    \item \textit{Attractiveness}: Is this description attractive? One of the goal of product description is to attract customer.
    \item \textit{Grammar}: Is the grammar of this description correct? Descriptions with low fluency decrease the quality of product description.
\end{compactitem}

In this set of experiments, we seek to determine if, and by how much,
the \textit{Extraction and Boosting} technique XB is able to improve the accuracy of the descriptive text and also not lose attractiveness and fluency. 
We randomly select $150$ product images and titles from FACAD and run \mantis{} alone and \mantis{} with our extracting and boosting technique \mantis\_XB to produce product descriptions. 
Human participants read the descriptions along with the product titles and images.
For the above three questions, participants answered whether the description met the criteria or not.

The results are shown in Figure~\ref{fig:eb_human}. 
Extracting the product attributes from images and then boosting the probability of the tokens which share high similarity with extracted attributes at the decoding stage of sequence generation improve the accuracy significantly. 
At the same time, it retains comparably attractiveness and fluency.

Our quantitative analysis reports model performance with the most commonly used natural language generation metrics, including BLEU \cite{papineni2002bleu}, METEOR\cite{denkowski2014meteor}, and ROUGE-L \cite{lin2004rouge} scores. Higher BLEU, METEOR, and ROUGE-L score indicate higher similarity with ground-truth descriptions.

Table~\ref{tab:auto_1} gives the metric scores for generated descriptions by \mantis\_XB and \mantis.
\mantis{} is trained by minimizing the difference between ground-truth description and generation, and thus, compared with \mantis\_XB, \mantis{} performs better in the dimensions of similarity with the ground-truth.
\mantis{} achieves a higher score in BLEU metrics.
However, \mantis{} is not statistically significantly better than \mantis\_XB in Rouge-L or METEOR.

We can conclude that after enforcing the satisfaction of given lexical constraints by controlling the decoding stage of sequence generation, accuracy of generated descriptions is significantly improved at no expense of other dimensions including attractiveness, grammar, and similarity with ground-truth descriptions.

\subsection{Reward Model Evaluation}
\label{sec:rw_eva}

After establishing that method ''XB`` improves the accuracy of the description, we train a reward model to predict whether the description will be preferred by humans. 
The architecture is illustrated in Section~\ref{sec:reward}.
We first train our reward model on FACAD to predict whether the description is ground-truth.
For each product, we label the ground-truth description as $0$ and the negatively sampled description from the same category as $1$.
Then we continue training our reward model on labeled generated descriptions in Section~\ref{sec:eb_eva} to predict human preference, given product title, images, attributes (optional) and descriptions.
We refer to these labeled generated descriptions as simply  ``generation''.
Table~\ref{tab:ex_rl} shows examples of labeled generated descriptions.
Human preferred descriptions are labeled as $0$ and descriptions contradicting their images or extracted attributes are labelled as $1$.

Table~\ref{tab:reward} shows the accuracy of reward model. 
We consider whether to use attribute modality of reward model---``w/ attr.'' and ``w/o' attr.' indicates applying and not applying attribute modality in reward model training, respectively.
We evaluate our reward model on the test set of generated descriptions ``generation'' in Section~\ref{sec:eb_eva} and FACAD.
With attribute modality, reward model is able to predict human preference with higher accuracy on generated descriptions and on sentences directly from FACAD.
We conclude that reward model is able to be used to predict human preference and produce produce reward and punishment values in Equation~\ref{eq:penalty}.

\subsection{Finetuning Evaluation}
\label{sec:rl_eva}

After training the reward model we assess whether the finetuned language model improves the accuracy of generated description. 
To this end, the reward model (Section~\ref{sec:rw_eva}) is used as a classifier to train \mantis{} using the policy gradient reinforcement learning technique in Section~\ref{sec:finetune}.
We compare the performance of the following three models on the metrics of accuracy, attractiveness and grammar (as in Section~\ref{sec:eb_eva}).
\begin{compactitem}
    \item \textbf{\mantis} \cite{jain2021multimodal} is a multimodal language model trained on FACAD.
    \item \textbf{\mantis\_RL}. Finetune \mantis{} with reward model using RL technique\cite{peng2020reducing}.
    \item \textbf{\mantis\_RL\_XB}. Filter generated descriptions of \mantis\_RL with XB technique.
\end{compactitem}

\begin{figure}[t]
    \centering
    \includegraphics[width=\linewidth]{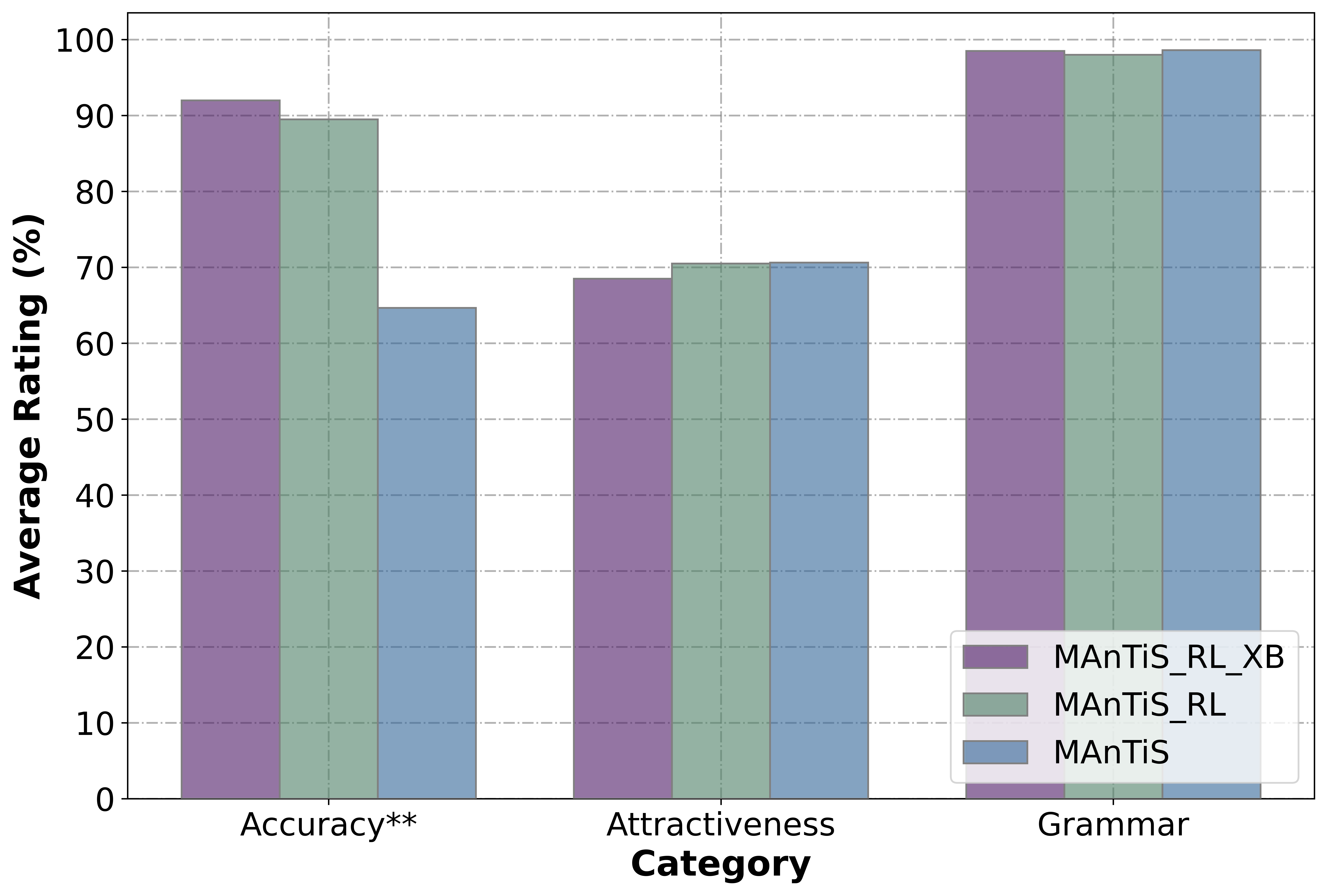}
    \caption{Human evaluation results on showing average rating of generated descriptions, $\ast\ast$ indicates $p < 0.01$.}
    \label{fig:rl_human}
\end{figure}

Figure~\ref{fig:rl_human} shows the result of human evaluation. 
After finetuning with reinforcement based on the reward model, accuracy of
generated description significantly increases.
Percentage of correct grammar and attractiveness remains steady after fine-tuning.
This indicates that the language models successfully learn human preference, according to the reward model, after the finetuning process completes. 
After applying the XB technique on \mantis\_RL, the accuracy of the descriptions is improved, but not statistically significant, indicating that the reinforcement learning process has successfully generalized the XB technique into the language model. 
In real-world applications of \oursys, description generation without control at the decoding stage XB could reduce computation cost.

\section{Conclusion}
Multimodal conditional language models are now widely used to produce image descriptions. However, all text generative models suffer from similar issues of quality control. \oursys{} provides a way to effectively address both obvious counterfactual problems as well as more subtle and idiosyncratic problems like overly-specific claims, incorrect tones, or other undesired phrases. Our framework can train any language model to generate human-preferred descriptions containing fewer errors by applying constraints at the decoding stage of generation and fintuning with a data-efficient policy-gradient method.

We see this work as a first step toward gaining greater levels of control on these powerful models, and a practical tool for decreasing the potential for unintended or unacceptable generations.

A thorough experimental study shows the proposed framework produces significantly more accurate descriptions compared to the strong baseline. These results demonstrate that \oursys{} trains multimodal language models to produce descriptive text with higher accuracy and better quality in general.

We believe \oursys{} can provide utility in many other domains as it is both flexible --- the reward model may be retrained to other preferences, as well as adaptable --- the system actively improves language models with additional feedback.

\section{Broader Impact}
Our system faces the same potential pitfalls as other contemporary language learning systems.
It is prone to echoing the biases present in existing datasets \cite{sheng2019woman} and generating non-normative text i.e., in violation of social norms. Future work may enable real-world facts to be injected into the decoding process for the purposes of journalism or misinformation.


\bibliography{anthology,acl2020}
\bibliographystyle{acl_natbib}

\end{document}

%% file: tables/auto_1.tex
\begin{table}
\footnotesize
\setlength\tabcolsep{1pt} 
\centering
\begin{tabular}{c|l|l|l|l|l}
\toprule
\textbf{Model} & \textbf{B-2 $\uparrow$} & \textbf{B-3 $\uparrow$} & \textbf{B-4 $\uparrow$} & \textbf{Rouge-L $\uparrow$} & \textbf{METEOR $\uparrow$} \\
\midrule
\mantis\_EB  &.223 &.191 &.164 & .277 & .172\\
MAnTiS & .257*  &.228* & .201* & .283 &.198\\
\bottomrule
\end{tabular}
\caption{Result of generator performance scores.
B-2, B-3 and B-4 denote 2-gram, 3-gram and 4-gram BLEU scores. * indicates the difference between \mantis\_EB and \mantis{} is statistically significant at $p <0.05$.}
\label{tab:auto_1}
\end{table}

%% file: tables/example_rl.tex
\begin{table*}
\footnotesize
\renewcommand{\arraystretch}{3}
{\begin{tabu}{p{0.08\textwidth} | p{0.25\textwidth}| p{0.5\textwidth}| p{0.05\textwidth}}
\toprule
\textbf{Image}  & \textbf{Title} & \textbf{Generated Description} & \textbf{Label} \\
\midrule
\multirow{2}{*}{\includegraphics[height=60pt,width=40pt]{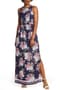}} 
& \multirow{2}{*}{Floral Print Maxi Dress}
& \multicolumn{1}{m{0.5\textwidth}|}{A ruffled high/low hem provides an airy update for a polka dot dress overlaid with a bright floral motif.} 
& 0
\\
\cline{3-4}
&&\multicolumn{1}{m{0.5\textwidth}|}{A ruffled hem and \textit{cap sleeves} add pretty finishing touches to a maxi dress in a vivacious floral print.}
& 1\\
\hline
\multirow{2}{*}[-1mm]{\includegraphics[height=60pt,width=40pt]{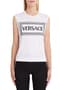}} 
& \multirow{2}{*}[-1mm]{Logo Tank}
& \multicolumn{1}{m{0.5\textwidth}|}{This logo tank gives any look a dose of rock 'n' roll cool.} 
& 0\\\cline{3-4}
&
& \multicolumn{1}{m{0.5\textwidth}|}{This logo-embroidered tank in \uwave{lightweight cotton} jersey is a casual homage to founder \uwave{Thomas Burberry}.}
& \multirow{1}{*}[-1.5mm]{1}\\
\bottomrule
\end{tabu}}
\caption{Example of generated description with human labels, where $0$ indicates \textit{human preferred} and $1$ indicates \textit{not preferred}. 
There are two types of not suitable descriptions, conflicting with image (in \textit{italic}) and over-specific information (\uwave{underlined}).
}
\label{tab:ex_rl}
\end{table*}

%% file: tables/reward.tex
\begin{table}
\footnotesize
\setlength\tabcolsep{6pt} 
\centering
\begin{tabular}{c|l|l}
\toprule
\textbf{Dataset} & \textbf{w/o attr. \% $\uparrow$} & \textbf{w/ attr. \% $\uparrow$}\\
\midrule
generation & 79.2 & \textbf{83.0} \\
FACAD & 70.6 & \textbf{72.3} \\
\bottomrule
\end{tabular}
\caption{Accuracy of reward model. ``w/o attr.'' and ``w/ attr.'' indicate training reward model without or with attribute modalities. 
}
\label{tab:reward}
\end{table}